\documentclass[12pt]{article}
\usepackage{scicite}
\usepackage[pdftex]{graphicx}
\usepackage{titling}
\usepackage{hyperref}
\usepackage{color,soul}
\usepackage[normalem]{ulem}               

\title{Learning as the Unsupervised Alignment of Conceptual Systems}
\author{
Brett D. Roads,$^{1\ast}$ Bradley C. Love$^{1}$\\
\\
\normalsize{$^{1}$Department of Experimental Psychology, University College London,}\\
\normalsize{WC1H 0AP, London, UK}\\
\\
\\
\normalsize{$^\ast$To whom correspondence should be addressed; E-mail:  b.roads@ucl.ac.uk.}
}

\date{}

\begin{document}

\maketitle

\paragraph{One Sentence Summary:} The meaning of concepts resides in relationships across encompassing systems that each provide a window on a shared reality.

\begin{abstract}
Concept induction requires the extraction and naming of concepts from noisy perceptual experience. For supervised approaches, as the number of concepts grows, so does the number of required training examples. Philosophers, psychologists, and computer scientists, have long recognized that children can learn to label objects without being explicitly taught. In a series of computational experiments, we highlight how information in the environment can be used to build and align conceptual systems. Unlike supervised learning, the learning problem becomes easier the more concepts and systems there are to master. The key insight is that each concept has a unique signature within one conceptual system (e.g., images) that is recapitulated in other systems (e.g., text or audio). As predicted, children's early concepts form readily aligned systems.
\end{abstract}

A typical person can correctly recognize and name thousands of objects. By 24 months, children already exhibit an average vocabulary of 200-300 words \cite{Fenson_etal_1994}. However, it remains unclear what mechanism makes this feat possible. Here, we conduct an information analysis and demonstrate that it is theoretically possible to learn the labels for objects through purely unsupervised means. Our key insight is that objects embedded within a conceptual system (e.g., text, audio, images) have a unique signature that allows for entire conceptual systems to be aligned (e.g., images with text) in an unsupervised fashion.

A common assumption is that some degree of explicit instruction is necessary for word learning. For example, a child might be told that a particular object is called a compass, or by reading a caption in a book, learn that a particular photograph depicts a ladybug. However, as V. W. O. Quine argued, even \emph{supervised} instruction contains a substantial amount of ambiguity \cite{Quine_1960}. If someone utters the word \emph{gavagai} while pointing to a rabbit, the word may refer to the whole animal, its long ears, the color of its fur, or the grass it's eating. Quine suggested that meaning may derive from something's place within a conceptual system. The meaning of gavagai could include all of these attributes as well as more macroscopic relationships such as the fact that rabbits are prey for other animals. Across multiple supervised learning episodes, it is possible for an individual to extrapolate the appropriate meaning of gavagai \cite{McMurray_etal_2012,Yu_Smith_2012}. However, a long-standing challenge of both cognitive science and machine learning is understanding how humans manage to learn concepts with relatively little supervised instruction.

Although the world is a noisy, bustling place with an indefinite number of learnable concepts, it is also trellised with statistical regularities. Given appropriate learning mechanisms, an agent can discover these statistical regularities through \emph{unsupervised} learning \cite{Bell_Sejnowski_1995,Chambers_etal_2003,Olshausen_Field_1996,Younger_Cohen_1986}. Unsupervised learning algorithms provide a means to construct rich feature representations--or \emph{embeddings}--of the corresponding inputs that capture meaningful semantic relationships  \cite{Caron_etal_2018,Pennington_etal_2014}. For example, an unsupervised learning system working with text documents would place cats and dogs near one another within a multidimensional embedding space because cats and dogs appear in similar linguistic contexts. However, such unsupervised approaches are siloed in that insights from one system (e.g., text) do not transfer to another (e.g., images). In contrast, human memory and semantic knowledge does appear to link the distributional statistics of different systems \cite{Tyler_Moss_2001}. Amodal semantic convergence zones in the anterior temporal lobe, in particular in perirhinal cortex, combine information across different sources \cite{Martin_etal_2018,OpDeBeeck_etal_2019}. Traditional unsupervised learning fails to address Quine's challenge or these observations from cognitive science.

One way to address Quine's challenge is to exploit temporal correlations across systems. At the broadest level, there are systematic correspondences across modalities, like that larger objects tend to generate lower-pitched sounds, and these relationship affect people's perceptual judgments \cite{Marks_1978}. Temporal correlations can be exploited by unsupervised techniques in order to link different sensations \cite{deSa_Ballard_1998}. These correlations also exist between language and the world and substantial work has investigated cross-situational word learning \cite{Fazly_Alishahi_Stevenson_2010,Goodman_Tenenbaum_Black_2008,Smith_Yu_2008}. A number of machine learning approaches encourage similar links by co-presenting multi-modal stimuli during training \cite{Kiela_Bottou_2014, Lazaridou_etal_2015, Ngiam_etal_2011, Ororbia_etal_2019}.

Although it is clear that humans can leverage temporal correlations, one question is whether unsupervised learning occurs in the absence of such temporal relationships. This important topic is relatively unexplored, although there is suggestive evidence that people can infer such linkages. For example, the fact that congenitally blind people come to organise semantic information in a similar fashion to sighted people, including visual information, is suggestive that information across modalities can be integrated asynchronously \cite{Lewis_etal_2019}. Establishing the value of such linkages, and subsequently developing algorithms to exploit these correspondences, would advance our understanding of both human and machine intelligence.

\begin{figure}
\centering
\includegraphics[width=1\linewidth]{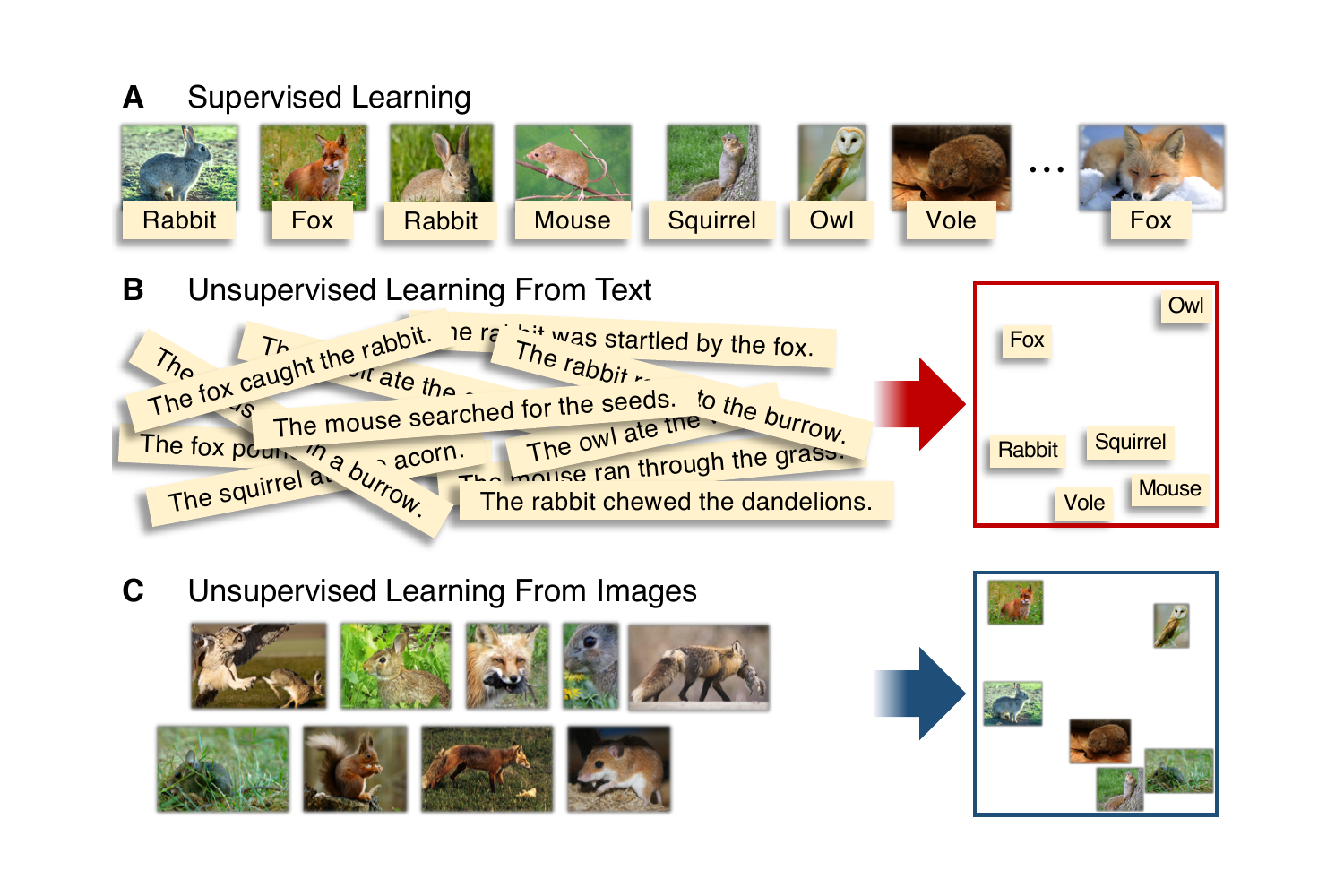}
\caption{Different modes of learning. \textbf{(A)} In supervised learning, feature representations are learned by co-presenting both an image and a label. Alternatively, unsupervised feature representations (i.e., embeddings) can be learned from \textbf{(B)} text and \textbf{(C)} images. The two unsupervised embeddings capture two viewpoints of the same underlying reality and therefore exhibit similar structure. Images shown here are from the public domain (Wikimedia Commons).}\label{fig:conceptual_example}
\end{figure}

Arguably, supervised learning is so powerful because it explicitly links distinct conceptual systems (e.g., images and words). In this work, we tap into this power by linking multiple embeddings in an unsupervised manner. In order to solve Quine's problem, we align a system of word labels, a system of visual semantics, and a system of audio semantics that all refer to the same underlying reality and therefore have related structure that can be discovered by unsupervised means (Figure~\ref{fig:conceptual_example}). We provide a computational-level \cite{Marr_1982} analysis that demonstrates how this process works.

Different sources of input should produce similar conceptual systems because sources are different viewpoints of the same underlying reality. For example, the concepts of \emph{grass}, \emph{rabbit}, \emph{mouse}, \emph{fox}, and \emph{owl} are likely to have similar co-occurrence statistics in visual media (images and videos) and communicative media (text and speech). In other words, functionally similar things tend to look alike, and we tend to talk in similar ways about things that are alike. If structural idiosyncrasies present in one embedding are qualitatively mirrored in the other embedding, then it is possible to align the two conceptual systems. In machine learning, a number of techniques referred to as \emph{manifold alignment} exploit similar assumptions in order to identify mappings between different conceptual systems \cite{Amodio_Krishnaswamy_2018,Ham_etal_2005,Wang_Mahadevan_2008,Wang_Mahadevan_2011}.

\begin{figure}
\centering
\includegraphics[width=1\linewidth]{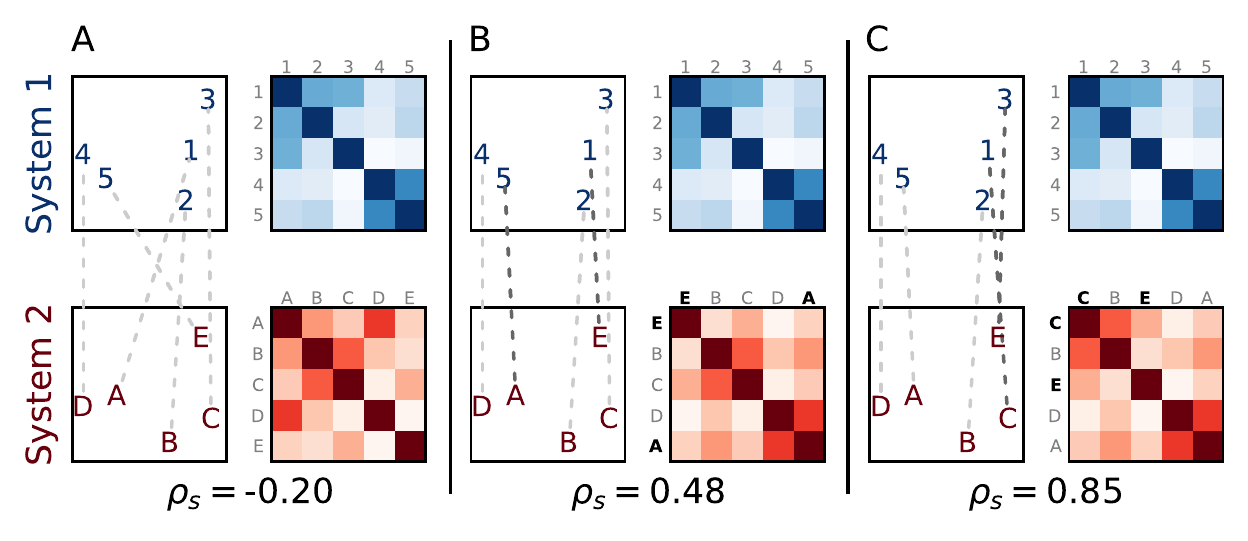}
\caption{Unsupervised linking of conceptual systems via conceptual alignment. A simplified scenario illustrating how unsupervised conceptual alignment can be used to identify a conceptual mapping between two systems. \textbf{(A)} A two-dimensional embedding of two distinct systems. Initially the mapping between the two systems is random (dashed gray lines) and the corresponding similarity matrices of the two embeddings look very different and have a low alignment correlation (i.e., Spearman correlation). \textbf{(B)} By swapping the mapping between concepts A and E, the alignment correlation is improved. \textbf{(C)} After making a second swap between concepts C and E, the best structural correspondence between the two embeddings has been found.}
\label{fig:alignment_cartoon}
\end{figure}

Aligning conceptual systems provides a means for an agent to continuously harvest information from everyday experience. Unlike supervised visual category learning--which requires images to be jointly presented with a label--conceptual alignment permits a learner to view many images without labels and many labels without images. By maximizing the conceptual alignment between the image-based and label-based embeddings, a mapping can be constructed between the two conceptual systems (Figure~\ref{fig:alignment_cartoon}). In the case of visual and speech input, identifying the \emph{correct mapping} would enable an agent to infer the correct verbal label for a visual stimulus, in a completely unsupervised manner.

Here, we align two (or more) unsupervised embedding spaces by creating a similarity matrix for each system and consider mappings between the systems. The similarity matrix captures the relational structure within each system. A good mapping or alignment reveals a second-order isomorphism between the systems \cite{Shepard_Chipman_1970}.

An \emph{alignment correlation} can be computed as the Spearman correlation between the upper diagonal portion of the two similarity matrices, where the mapping determines the order of concepts in the matrices. A correct mapping will link each concept in one system (e.g., the image of a dog) with the corresponding concept in the other system (e.g., the word ``dog''). Given the concept intersection $\mathcal{C}$ between two systems, there are $|\mathcal{C}|!$ potential one-to-one mappings, of which, only one is the correct mapping. Mapping concepts in one system, to those in another system that play a similar role, will increase the alignment correlation.

In our computational studies, we have a ground truth view on the system alignment, so can measure the objective quality of a particular mapping by its accuracy, i.e., the number of concepts that are correctly mapped from one system to another. For unsupervised system alignment to be useful, alignment correlations should positively correlate with objective accuracy. Furthermore, one would expect the correct mapping to have a high alignment correlation relative to the majority of other mappings.

\section*{Results}

\begin{figure}
\centering
\includegraphics[width=1.\linewidth]{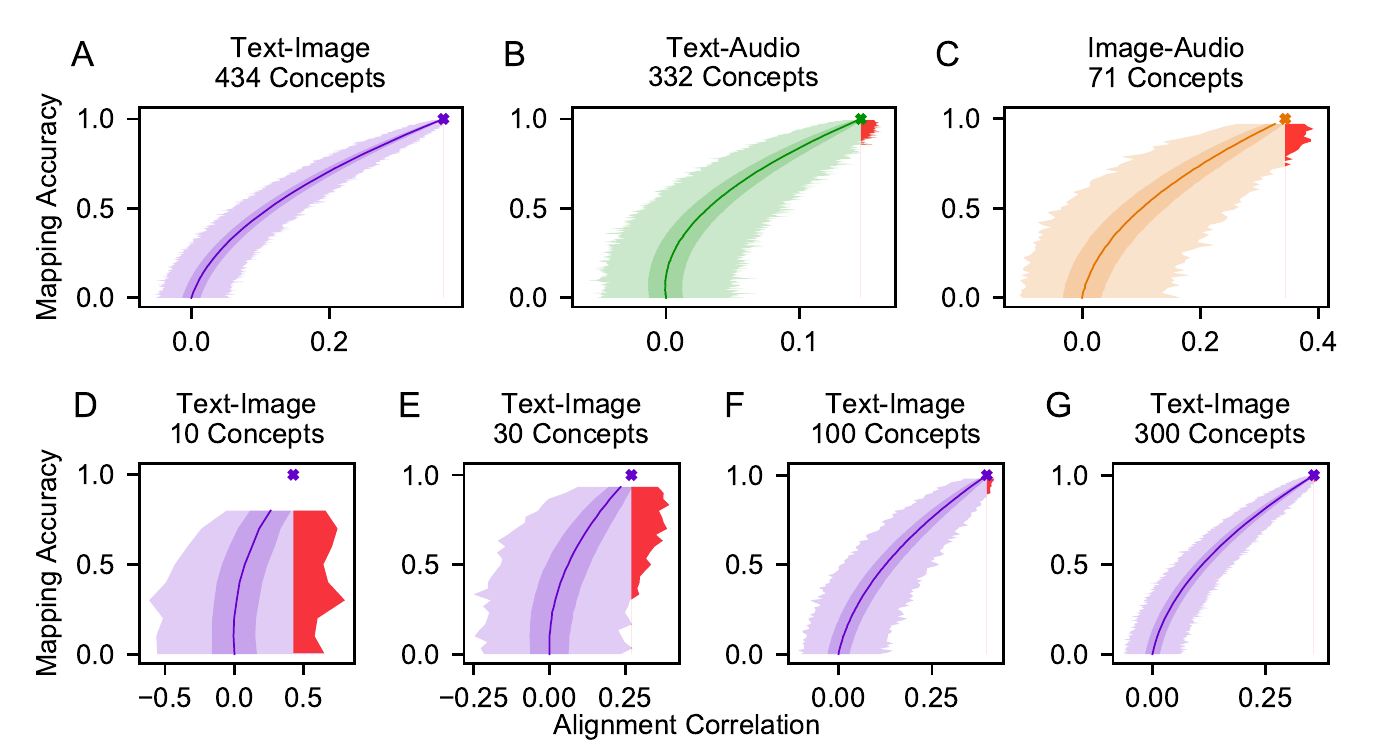}
\caption{Alignment correlation versus mapping accuracy for three real-world datasets. Each scenario revealed the distribution of alignment correlations conditioned on mapping accuracy (i.e., the proportion of correctly matched concepts). When the number of concepts is large, there was a strong correlation when aligning \textbf{(A)} text and images, \textbf{(B)} text and audio, and \textbf{(C)} images and audio. Focusing on alignment between text and images, using fewer concepts resulted in an increasingly weaker correlation \textbf{(D-G)}. Each plot shows the mean alignment correlation (dark line), a one standard deviation envelope (dark shading), the minimum/maximum envelope (light shading), and the alignment correlation of the correct mapping (dark X). The sample statistics (mean, standard deviation, minimum, and maximum) for each horizontal slice are based on 10,000 randomly sampled mappings or all unique mappings. The red regions indicate \emph{misleading mappings}--imperfect mappings that have a higher alignment correlation than the correct mapping. The red regions highlight how maximizing alignment correlation does not guarantee the best mapping.}
\label{fig:fig_3}
\end{figure}

We found that alignment correlations positively correlated with mapping accuracy across a variety of scenarios (Figure~\ref{fig:fig_3}A-C). The three conceptual systems were derived from a Common Crawl text corpus \cite{Pennington_etal_2014}, the Open Images dataset \cite{OpenImages}, and the AudioSet dataset \cite{Gemmeke_etal_2017}. For simplicity, these datasets are referred to as the text, image, and audio datasets. Corresponding unsupervised embeddings for each dataset were created using the GloVe algorithm \cite{Pennington_etal_2014}.

Each scenario was created by taking the concept intersection between two datasets and randomly sampling mappings between the two systems. Mappings were conditionally sampled based on their accuracy. For each level of accuracy (e.g., three incorrectly mapped concepts), 10,000 unique mappings were sampled and their alignment correlations computed. If there were less than 10,000 unique mappings then all available mappings were used. Conditional sampling was necessary since there are substantially more ways to assemble low-accuracy mappings than high-accuracy mappings. The Spearman correlation between the mapping accuracy and conditionally-sampled mapping alignment correlation was $\rho=.99$ ($p<.01$) for the text-image, $\rho=.92$ ($p<.01$) for the text-audio, and $\rho=.92$ ($p<.01$) for the image-audio scenarios.

A concept's signature of its place within a conceptual system is richer the bigger the system. The correlation between alignment correlations and mapping accuracy increases as the number of concepts increases (Figure~\ref{fig:fig_3}D-G). Smaller scenarios are constructed by using a subset of the original text and image concepts. In the same manner as before, up to 10,000 mappings are sampled for each level of mapping accuracy and their corresponding alignment correlations computed. For 10, 30, 100 and 300 concepts, the Spearman correlation is $\rho=.16$ ($p<.01$), $\rho=.67$ ($p<.01$), $\rho=.96$ ($p<.01$), and $\rho=.98$ ($p<.01$) respectively. 

A concept's signature is weaker in scenarios with fewer concepts. As a thought experiment, in the extreme of only two concepts, there would be no unique signature. The more concepts there are, the greater the chance that concepts will be disambiguated from one another by virtue of each concept's similarity relations within the system. As shown by the uncertainty envelope in Figure~\ref{fig:fig_3}D-G, the smaller the system, the more likely one is to happen upon a imperfect solution that has a misleadingly high alignment correlation. \emph{Misleading mappings} (i.e., imperfect mappings with a higher alignment correlation than the correct mapping) arise due to structural deviations between different systems. For example, imagine that the concepts \emph{fox} and \emph{rabbit} were switched in word-based embedding of Figure~\ref{fig:conceptual_example}. Maximizing alignment correlation would erroneously map the word ``fox'' to an image of rabbit.

To quantify the prevalence of misleading mappings, we introduce the \emph{alignment strength} measure. When there are no misleading mappings, the alignment strength is 1. When all incorrect mappings are misleading, alignment strength is 0. The corresponding alignment strengths of the previously discussed scenarios are plotted in Figure~\ref{fig:fig_s1}. In agreement with the previous correlation analysis, alignment strength is low for few-concept scenarios and high for many-concept scenarios.

So far we have considered mapping accuracy in an all-or-none fashion. However, some incorrect mappings are intuitively worse than others. For example, mapping the concept \emph{pear} to \emph{violin} seems qualitatively worse than mapping \emph{pear} to \emph{apple}. Focusing on three groupings of concepts (birds, musical instruments, fruits), we consider mappings where two concepts are misaligned. In one case, the misaligned concepts come from the same grouping (e.g., both fruits). In the other case, the misaligned concepts come from different groupings. By considering all within- and across-group pair-wise errors, we compute the percentage by which the alignment correlation becomes worse compared to a perfect mapping. On average, a within-group misalignment reduces the alignment correlation by 0.16\% and an across-group misalignment by 1.31\%, which is roughly an eight-fold effect on alignment correlation for near versus far errors.

\begin{figure}
\centering
\includegraphics[width=1.\linewidth]{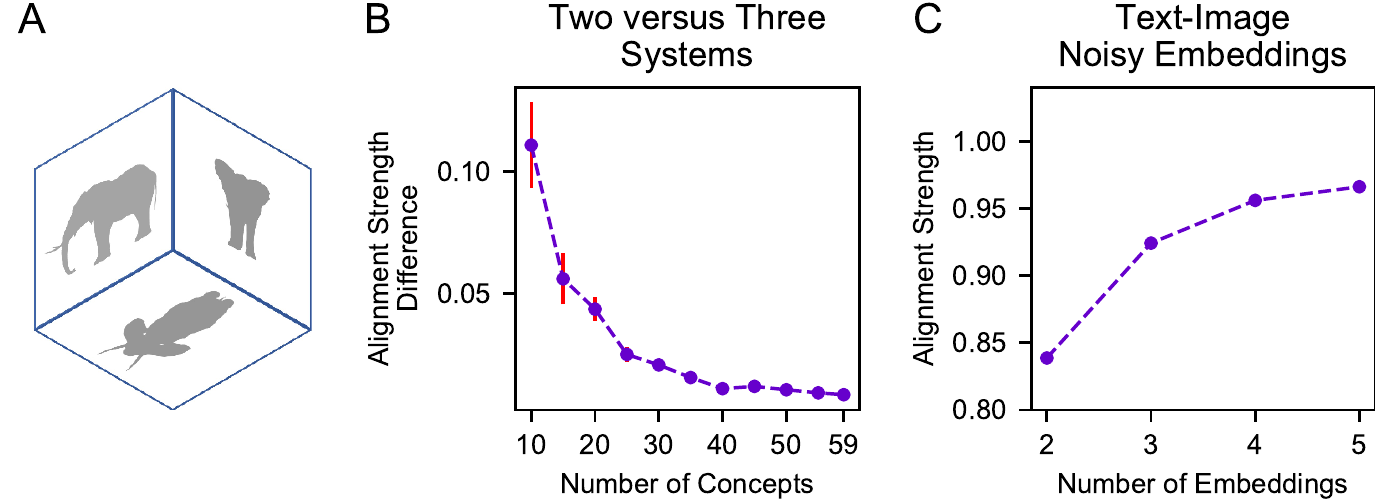}
\caption{The impact of multiple conceptual systems on alignment strength. \textbf{(A)} Much like how multiple visual viewpoints of the same object can aid understanding, so can integrating information from multiple conceptual systems.  \textbf{(B)} Results for aligning three different conceptual systems. The line shows the alignment strength of three systems (text-image-audio) less the alignment strength of two systems (text-image). Each data point represents the average alignment strength obtained from 50 randomly sampled subsets. Red error-bars indicate standard error of the mean. \textbf{(C)} Scenarios where conceptual alignment is performed between a text-based embedding and multiple noisy image-based embeddings. The horizontal axis indicates the total number of embeddings used during conceptual alignment. Each data point is obtained by computing the mean alignment strength for 10 different scenarios. Red error-bars indicate standard error of the mean.}
\label{fig:fig_4}
\end{figure}

The previous results examine conceptual alignment by linking two conceptual systems. In principle, conceptual alignment can be performed with more than two systems. Each conceptual system can be likened to a different viewpoint of the same reality, where additional viewpoints improve an agent's ability to infer the correspondence between the various perspectives (Figure~\ref{fig:fig_4}A). Adding a third system yields a higher alignment strength compared to using only two systems (Figure~\ref{fig:fig_4}B). When leveraging the structural idiosyncrasies of three systems, the correct mapping becomes more competitive relative to the incorrect mappings. Conducting adjusted t-tests of independent samples using the Holm-Bonferroni method ($\alpha=.05$) show a significant improvement of alignment strength for all subset sizes. The smallest subset size (10 concepts) exhibits the largest improvement, with the three-system alignment strength ($M=0.61$, $SD=0.08$) larger than the two-system alignment strength ($M=0.50$, $SD=0.10$), $t(49)=6.27$, $p<0.001$. The remaining subset sizes exhibit a significant, but decreasing improvement over the two-system alignment strength. For the largest subset size (59 concepts), the three-system alignment strength ($M=0.954$, $SD=0.001$) is only marginally better than the two-system alignment strength ($M=0.945$, $SD=0.002$), $t(49)=31.43$, $p<0.001$.

A complementary method to evaluate the benefit of using more than two conceptual systems is to consider a scenario in which some of the embeddings have been corrupted by noise. Given that our individual experiences are noisy, an individual's mental embedding of a system is also likely to be noisy. After adding a sufficient amount of noise, performing conceptual alignment between a text embedding and a noisy image embedding reduces the alignment strength from approximately .99 to .85. The alignment strength is partially restored by including an increasing number of noisy image embeddings during conceptual alignment (Figure~\ref{fig:fig_4}C). After including five different embeddings, the alignment strength was restored to approximately .95. Analogously, people may rely on multiple senses and information sources when forming an integrated semantic representation.

Given the difficulty of aligning conceptual systems when there are few concepts, it is interesting to consider how infants and children accomplish this task. One possibility is that the early concepts children acquire are the ones that form conceptual systems that can be aligned without supervision. We evaluate this possibility by comparing alignment strength for a random subset of concepts to a subset of the earliest acquired concepts \cite{Kuperman_etal_2012}. As predicted, incorporating age-of-acquisition (AoA) for different concepts improves alignment strength, particularly when the systems consist of few concepts (Figure~\ref{fig:fig_s2}). Adjusted t-tests using the Holm-Bonferroni method ($\alpha=.05$) shows the largest boost for the smallest subset size, where the AoA-constrained alignment strength ($M=0.65$, $SD=0.14$) is substantially greater than the unconstrained alignment strength ($M=0.53$, $SD=0.09$) $t(19)=3.10$, $p<0.01$.

\section*{Discussion}

Arguably, much of the power of supervised learning comes from providing direct links between distinct conceptual systems. Most unsupervised approaches learn conceptual systems in a siloed fashion, failing to bridge different systems. We showed that a strong signal exists that can guide unsupervised alignment to solve this problem. Each concept has a unique signature within one conceptual system (e.g., images) that is mirrored in other systems (e.g., text and audio). While an agent may build distinct conceptual systems from different sensory modalities, the sources of experience originate from a shared reality. This enables links to be made between different systems to potentially support low-shot or zero-shot learning. Rather than mastering isolated systems, the learning problem can be characterized as aligning entire conceptual systems.

In keeping with the system alignment perspective, as the number of concepts increases, the correlation between mapping accuracy and alignment correlation also increases. This suggests that including additional concepts creates a richer, more distinctive, relational structure, which in turn favors mappings that are mostly correct. In scenarios involving many concepts, the structural relationships are sufficiently unique that alignment correlations could serve as a strong prior for learning concept mappings, reducing the need for supervised learning. Furthermore, alignment correlation favors sensible mistakes, where within-group misalignments (e.g., pear to apple) have a better score than across-group misalignments (e.g., pear to violin. In each embedding space, structural relationships among members act as distinguishing landmarks that an agent can use to align different systems. It is possible that this pattern contributes to the vocabulary spurt exhibited by some children \cite{Goldfield_Reznick_1990}.

Conceptual alignment struggles in scenarios involving very few concepts. However, alignment strength can be boosted by linking more than two conceptual systems. Interestingly, alignment in low-concept scenarios can also be increased by restricting analysis to the set of words acquired earliest in life. Difficult-to-align systems contain concepts that are equally similar to all other concepts. When all concepts are equally similar, there is no structure in the similarity relationships and no way to map concepts in one system to another system, since there is no way to resolve ambiguity. In contrast, easy-to-align systems exhibit structural relationships that make them much more distinctive. It is important to note that an early acquired word like ``Toothbrush'' is not always easy to align. Rather, the concept of toothbrush in the context of other early-acquisition words, creates a system that engenders a unique signature for toothbrush.

An interesting possibility is that certain sets of words are more likely to be acquired early because they form distinctive structural relationships and are therefore easier to map. Alternatively, caregivers may have an implicit understanding of these relationships and curate their interactions to promote the learning of these less ambiguous systems \cite{Samuelson_2002}. Relatedly, it is conceivable that more frequently experienced concepts are more readily alignable, but it seems equally plausible that more alignable concepts are experienced more frequently. The fact that children tend to produce basic level nouns first \cite{Mervis_1987}, might be partially explained by the distinctive structural relationships of nouns \cite{Jones_etal_1991,Samuelson_Smith_1999}. A drive to align conceptual systems may also help explain why information from multiple modalities can facilitate learning in infants \cite{Frank_etal_2009}.

These initial results open a host of possibilities and future challenges. One basic question is what must be assumed to successfully align conceptual systems. For example, a majority of our analysis involved embedding spaces that relied on annotated data, rather than raw images. Additional analyses were conducted using embeddings derived solely from pixel-level information use the DeepCluster algorithm \cite{Caron_etal_2018}. Amazingly, a relatively high alignment strength can still be obtained when performing conceptual alignment between pixel-based embeddings and text-based embeddings, a feat that would not have been possible 10 years ago. Future advances in extracting such spaces through unsupervised means should lead to improved alignments.

While our aim was to demonstrate that information on correct mappings is present across unsupervised embedding spaces, one challenge for future work is discovering how to efficiently search through the vast space of possible alignments to discover a suitable mapping. We predict that this challenge will be addressed by search algorithms that leverage basic constraints on cognition \cite{Spelke_Kinzler_2007,Ullman_etal_2012} to efficiently approximate the optimal solution, much like how analogy models that align individual concepts have progressed \cite{Gentner_1983,Holyoak_etal_1989,Larkey_Love_2003}.

\section*{Methods}

The primary objective of the study is to determine if different conceptual systems can be aligned in an unsupervised fashion. The secondary objectives of this study are to determine how the number of concepts influences alignment, how the number of conceptual systems influences alignment, and how alignment performance compares to human word acquisition. All of these objectives are pursued using a relatively algorithm-agnostic approach in order to best understand the theoretical properties of system alignment rather than the capabilities of a particular alignment algorithm. 

The study is organized into two sequentially-dependent stages. The first stage assembles conceptual systems from real-world datasets using two different embedding techniques. The second stage uses the assembled conceptual systems in order to achieve the research objectives. To achieve the study's research objectives, we use multiple real-world datasets in order to assemble distinct conceptual systems. The conceptual systems are assembled in an unsupervised fashion using two different embedding techniques. One technique leverages co-occurrence statistics and the Glove algorithm \cite{Pennington_etal_2014}. The second technique uses deep neural networks \cite{Caron_etal_2018}.

\subsection*{Assembling Conceptual Systems via Co-occurrence Embeddings}
Separate embeddings are derived by applying the GloVe algorithm to co-occurrence statistics collected from each domain. Co-occurrence statistics are tracked using a symmetric co-occurrence matrix $X$, where element $X_{ij}$ indicates the co-occurrence frequency of the $i$th and $j$th concept. Co-occurrence statistics are assembled in slightly different ways for each domain. Although the GloVe algorithm was originally designed to work with text corpora, it is well-suited to work with co-occurrence data derived from other sources. In it's original formulation, co-occurrence statistics are assembled using a sliding window that traverses the entire corpus. When words co-occur in a window, the corresponding element in the co-occurrence matrix is increased. The magnitude of the increment is modulated be the distance between the two co-occurring words. Words that are close together receive a larger increment than words that are far apart. In an analogous way, co-occurrence statistics can be assembled from other media such as images and audio. When two concepts co-occur in the same image or within the same audio file the corresponding element in the co-occurrence matrix is incremented. In this work, co-occurrence counts for images and audios are not weighted by a distance function.

Given a co-occurrence matrix, the GloVe algorithm then infers an embedding $W$. The embedding ($W$) is inferred by minimizing the following loss function
\begin{equation}
J = \sum_{i,j=1}^{V} f(X_{ij}) ( w_{i}^{T} \tilde{w}_{j} + b_i + \tilde{b}_j - \log{X_{ij}} )^{2},
\label{eqn:glove_cost}
\end{equation}
where $V$ indicates the number of unique concepts and the $b$'s are jointly inferred bias terms. The weighting function $f$ is given by:
\begin{equation}
f(x) =
\left\{\begin{array}{r@{}l@{\qquad}l}
    (\frac{x}{x_{\max}})^{\alpha} & &\textrm{if}\ x\leq x_{\max} \\[\jot]
    1 & &\textrm{otherwise},
  \end{array}\right.
\end{equation}
where $x_{\max}=100$ and $\alpha=.75$.

The text embedding used in this work is a publicly available embedding that has been pretrained \cite{Pennington_etal_2014}. The pretrained embedding produces 300 dimensional vectors for each word in the vocabulary set. Word co-occurrence statistics were derived from a Common Crawl corpus composed of 840 billion tokens and 2.2 million, cased vocabulary words.

The image-based co-occurrence statistics are derived from the publicly available Open Images V4 dataset \cite{OpenImages} (Boxes subset). The Open Images V4 dataset contains class annotations for approximately 9 million images. Each image has been annotated by a human, machine or both to indicate which of 19,995 classes are present. The majority of images contain multiple classes. Instead of using a window to determine co-occurrence, all classes present in a given image are treated as co-occurring. When the co-occurrence matrix is incremented, all co-occurrences are treated equally and incremented by 1. In this dataset, there is a mean of 3.8 classes per image (SD=2.5), with a maximum of 31 classes in an image. To infer an embedding from the co-occurrence matrix, we use the GloVe cost function (Equation~\ref{eqn:glove_cost} with the same free-parameter values for $x_{\max}$ and $\alpha$. To reduce the likelihood of overfitting, we assume a smaller dimensionality of 50.

The audio-based co-occurrence statistics are derived from the publicly available AudioSet dataset \cite{Gemmeke_etal_2017}. The AudioSet dataset contains approximately 2 million 10-second audio files drawn from YouTube videos. Each audio file has been human annotated to indicate the presence of 632 different event classes. The majority of audio files contain multiple classes. Like the image-based approach, classes that occur in an audio file are treated as co-occurring and treated equally regardless of location in the audio file. In this dataset, there is a mean of 2.0 classes per file (SD=1.2), with a maximum of 15 classes in a file. The parameters used to infer an image-based embedding were used to infer a 50-dimensional audio embedding.

Future work could consider the benefit of weighting image and audio co-occurrence using some form of distance function. For example, audio co-occurrence statistics could be incremented based on the temporal separation of events. Likewise, a spatial model might be used to weight co-occurrence in images.

These three embeddings provide the three core conceptual systems that are used in later analysis. For simplicity, these datasets are referred to as the text, image, and audio conceptual systems.

\subsection*{Assembling Conceptual Systems via Deep Neural Network Embeddings}
Pixel-level embeddings are obtained using a VGG-16 neural network \cite{Simonyan_2014} that has been pretrained using the DeepCluster approach, which yields a 4096 dimensional feature vector for each image \cite{Caron_etal_2018}. All images from the training set of the ImageNet dataset \cite{Deng_2009} were encoded using the pretrained model. Since there are many images for each concept, a conceptual system was assembled by randomly sampling one image-encoding for each class. In other words, one image from each concept is randomly selected to serve as the representative of that concept. By repeating this process 25 times, 25 different conceptual systems were assembled from the ImageNet dataset. Since this embedding technique leverages pixel-level information, this embedding is referred to as the pixel conceptual system.

\subsection*{Aligning Conceptual Systems}

The assembled conceptual systems are used to create a number of scenarios in order to evaluate unsupervised conceptual alignment. Three different classes of scenarios are considered: scenarios involving two systems, scenarios involving more than two systems, and scenarios that leverage age-of-acquisition information for words. Each scenario was created by taking the concept intersection between the included systems. Analysis focused on alignment correlation and alignment strength.

For each scenario, analysis is restricted to the concept intersection of all the systems involved. To determine an intersection between a set of domains, we first determined a single word label for every concept in each domain. The concepts derived from the Common Crawl text corpus required no additional processing since the embedding procedure was performed for single word tokens \cite{Pennington_etal_2014}. Single word labels were obtained for the OpenImages, AudioSet, and age-of-acquisition datasets by dropping all concepts described by more than one word. Single word concepts for the ImageNet dataset were obtained by manually coding the provided descriptions into single word tokens. Future work could expand the analysis by considering concepts describe by more than one word.

\subsection*{Alignment Strength}

Once the concept intersection has been determined, an alignment strength analysis uses multiple independent runs to guarantee that the results are representative of a general pattern. For each run, a random subset of concepts is used such that five concepts remain unused. For example, the intersection of the text, image, and audio dataset contains 64 concepts. For each run, 59 concepts are randomly drawn from the 64 concepts. Leaving out five concepts enforces a degree of variability across the different runs and confidence in the results.

\subsection*{Computing Mapping Accuracy and Alignment Correlation}

In scenarios where two systems are being aligned, accuracy is determined by the number of concepts that have been correctly mapped from one system to another. The alignment correlation is the Spearman correlation between the upper diagonal portion of the two similarity matrices, where the mapping determines the order of concepts in the matrices.

In scenarios aligning more than two systems, mapping accuracy is determined in an all-or-none fashion. Only concepts that have been correctly mapped across all systems are counted as correct. For example, in a three system scenario, matching \textit{dog word}, with \textit{dog audio}, with \textit{dog image} would count as one correct mapping. In contrast, matching the \textit{dog word}, with \textit{dog audio}, with \textit{cat image} would be treated as an incorrect mapping, with no partial credit given. To compute accuracy, the total number of correct mappings is divided by the the maximum number of possible correct mappings (i.e., the number of concepts).

Since correlations examine pairs of variables, we employ a slightly more complicated approach when computing alignment correlation for more than two systems. Given $N$ systems, there are $N$-choose-$2$ unique pairs of systems. For each of these system pairs, we compute the alignment correlation. To obtain a final composite alignment correlation, we take the mean of all pair-wise alignment correlations.

\paragraph{Data Availability:}
The source datasets used in this work are available for download from their corresponding sources. The ImageNet images are availabe from \url{http://www.image-net.org/}. The OpenImages V4 Boxes dataset is available from \url{https://storage.googleapis.com/openimages/web/download_v4.html}. The AudioSet dataset is available from \url{https://research.google.com/audioset/download.html}. The pre-trained GloVe embedding (Common Crawl 840B word tokens) is available from \url{https://nlp.stanford.edu/projects/glove/}. The age-of-acquisiton ratings are available from \url{http://crr.ugent.be/papers/AoA_ratings_Kuperman_et_al_BRM.zip}. The concept intersection for the analysis are permanently hosted at the OSF repository \url{https://osf.io/ndrmg/}.

\paragraph{Code Availability:}
The Python code used to perform the analysis in this work is permanently hosted at the OSF respository \url{https://osf.io/ndrmg/} and is licensed under the Apache License 2.0. The code repository for computing image embeddings using the DeepCluster algorithm is located at \url{https://github.com/facebookresearch/deepcluster}. DeepCluster is licensed under a Creative Commons Attribution-NonCommercial 4.0 International Public License.

\paragraph{Declarations:} The authors declare that they have no competing interests.

\paragraph{}Correspondence and requests for materials should be addressed to B.D.R.

\paragraph{Acknowledgements:} This work was supported by NIH Grant 1P01HD080679, Wellcome Trust Investigator Award WT106931MA, and Royal Society Wolfson Fellowship 183029 to B.C.L.

\paragraph{Author Contributions:} B.D.R. and B.C.L. designed the research. B.D.R. designed and implemented the analysis. B.D.R. and B.C.L. discussed all aspects of the implementation of the analysis and figures. B.D.R. and B.C.L. wrote the paper.


\setcounter{section}{1}
\renewcommand*{\thesection}{S~\arabic{section}}
\renewcommand*{\thesubsection}{\arabic{section}.\arabic{subsection}}
\setcounter{figure}{0}
\renewcommand\thefigure{S\arabic{figure}}

\title{Supplementary Materials for\\Learning as the Unsupervised Alignment of Conceptual Systems}
\author{
Authors: Brett D. Roads,$^{1\ast}$ Bradley C. Love$^{1}$\\
\\
\normalsize{Affiliations: $^{1}$Department of Experimental Psychology, University College London,}\\
\normalsize{WC1H 0AP, London, UK}\\
\\
\\
\normalsize{$^\ast$To whom correspondence should be addressed; E-mail:  b.roads@ucl.ac.uk.}
}
\maketitle

\subsubsection*{This PDF file includes:}
\begin{description}  
\item Figs. S1 to S2
\end{description}

\begin{figure}
\begin{center}
\includegraphics[width=1\linewidth]{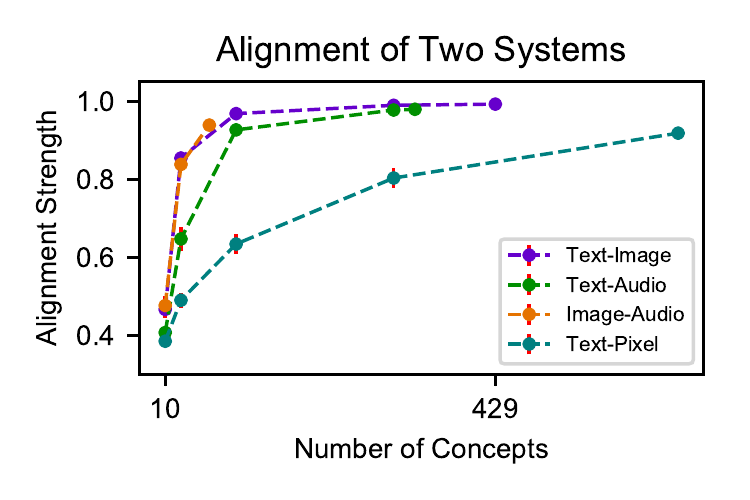}
\end{center}
\caption{The alignment strength of two-system alignment scenarios. The size of the subset is denoted by the horizontal axis and the average alignment strength is denoted by the vertical axis. Each data point represents the mean alignment strength of 50 independent runs. Each run uses a random subset of the specified number of concepts. In the case of the text-pixel scenarios, each data point represents the mean alignment strength of 25 different versions and 50 independent runs. Red error-bars indicate standard error of the mean.}
\label{fig:fig_s1}
\end{figure}

\begin{figure}
\begin{center}
\includegraphics[width=1.\linewidth]{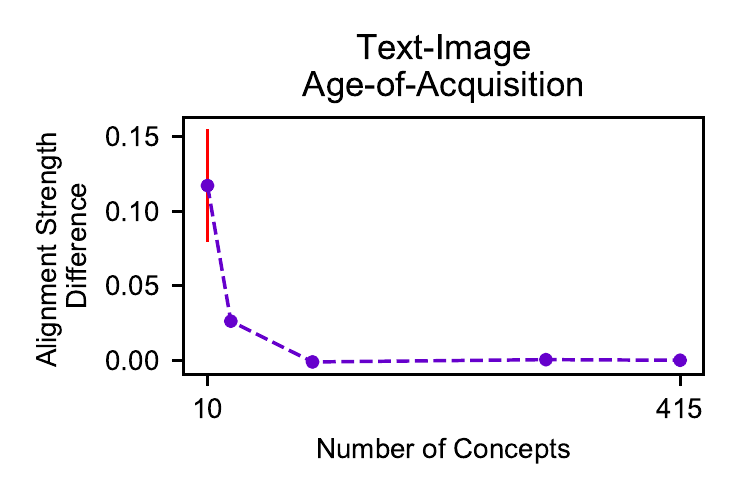}
\end{center}
\caption{Alignment strength using age-of-acquisition data. The difference between alignment strength for age-of-acquisition subsets and random subsets as a function of number of concepts used during alignment. Each data point represents the mean alignment strength difference of 50 randomly sampled subsets. Red error-bars indicate standard error of the mean. When age-of-acquisition data is taken into account. Subsets are created by selecting concepts that have the lowest age-of-acquisition for the a particular subset size.}
\label{fig:fig_s2}
\end{figure}

\end{document}